\documentclass[twocolumn,10pt]{IEEEtran}
\usepackage{indentfirst}
\usepackage{amsmath}
\usepackage{amsfonts}
\usepackage{amssymb}
\usepackage{cite}
\usepackage{times}
\usepackage{booktabs}
\usepackage[dvips]{graphicx}
\usepackage{hhline}
\usepackage{color}
\usepackage{multirow}
\usepackage{makecell}
\usepackage[table]{xcolor}
\usepackage{graphicx}
\usepackage{epstopdf}
\usepackage{algpseudocode}
\usepackage{algorithm}
\usepackage{adjustbox}
\graphicspath{{./plots_for_errata/} }
\usepackage{caption, subcaption}

\usepackage[absolute]{textpos}
\usepackage{everyshi}

\usepackage{url}

\usepackage[shortlabels]{enumitem}

\usepackage{color,soul}

\newcommand{\bi}{\begin{itemize}}
\newcommand{\ei}{\end{itemize}}

\newcommand{\be}{\begin{IEEEeqnarray}}
\newcommand{\ee}{\end{IEEEeqnarray}}

\newcommand{\commentLB}[1]{}
\usepackage[acronyms,nonumberlist,nopostdot,nomain,nogroupskip]{glossaries}
\newacronym{eeg}{EEG}{electroencephalography}
\newacronym{emg}{EMG}{electromyography}
\newacronym{eog}{EOG}{electroculography}
\newacronym{ecg}{ECG}{electrocardiography}
\newacronym{msc}{MSC}{magnitude squared coherence}
\newacronym{roi}{ROI}{regions of interest}
\newacronym{ml}{ML}{machine learning}
\newacronym{cv}{CV}{cross-validation}
\newacronym{rbf}{rbf}{radial basis function}
\newacronym{svm}{SVM}{Support Vector Machine}
\newacronym{ioth}{IoTH}{Internet-of-Things for healthcare}
\newacronym{fft}{FFT}{Fast Fourier Transform}
\newacronym{nn}{NN}{neural network}
\newacronym{qos}{QoS}{quality of service}
\newacronym{fesc}{FeSC}{Feature Selection with Consensus}

\usepackage{geometry}
\begin{document}

 \title{Feature selection for gesture recognition in Internet-of-Things for
   healthcare}

 \author{Giulia Cisotto$^{1,2}$, Martina Capuzzo$^{1}$, Anna V. Guglielmi$^{1}$, Andrea Zanella$^{1}$\\
   $^{1}$Dept. of Information Engineering, University of Padova, Padova, Italy\\
   $^{2}$Integrative Brain Imaging Center, National Center of Neurology and Psychiatry, Tokyo, Japan\\
   % via Gradenigo 6B, 35131 Padova, Italy \\
   email: \{cisottog, capuzzom, guglielm, zanella\}@dei.unipd.it }

 \date{}
 \maketitle
 \thispagestyle{empty} \pagestyle{empty}

 \begin{abstract}
   % \hl{MAX: 250 words} --> OK
   Internet of Things is rapidly spreading across several fields, including
   healthcare, posing relevant questions related to communication capabilities,
   energy efficiency and sensors unobtrusiveness. Particularly, in the context
   of recognition of gestures, e.g., grasping of different objects, brain and
   muscular activity could be simultaneously recorded via EEG and EMG,
   respectively, and analyzed to identify the gesture that is being
   accomplished, and the quality of its performance. This paper proposes a new
   algorithm that aims (i) to robustly extract the most relevant features to
   classify different grasping tasks, and (ii) to retain the natural meaning of
   the selected features. This, in turn, gives the opportunity to simplify the
   recording setup to minimize the data traffic over the communication network,
   including Internet, and provide physiologically significant features for
   medical interpretation. The algorithm robustness is ensured both by consensus
   clustering as a feature selection strategy, and by nested cross-validation
   scheme to evaluate its classification performance.
   %
%   Results show that the algorithm achieves a good classification performance
%   (accuracy over \hl{$0.9$}) with a limited number of physiologically relevant
%   features. In the future, generalization to other classes of biosignals could
%   represent a straightforward extension of this work to monitor other kinds of
%   activities.
   Although \gls{fesc} implements a very robust architecture for feature selection and classification,
   results are still negatively affected by the limited size of the dataset. In the future,
   further investigations could determine to what extent size could cause a drop in the performance
   of \gls{fesc} in this and other gesture recognition applications.
 \end{abstract}

% For PRE-PRINT on ArXiv:
\begin{textblock*}{17cm}(1.7cm, 0.5cm)
\noindent\scriptsize This paper will be published on \emph{2020 IEEE International Conference on Communications, Virtual Conference (Dublin, Ireland), June 2020.}\\
\textbf{Copyright Notice}: \textcopyright 2020 IEEE. Personal use of this material is permitted. Permission from IEEE must be obtained for all other uses, in any current or future media, including reprinting/republishing this material for advertising or promotional purposes, creating new collective works, for resale or redistribution to servers or lists, or reuse of any copyrighted component of this work in other works.
\end{textblock*}


 \glsresetall

 %% INTRODUCTION
 \section{Introduction}
 % Intro + related works

 %% SCENARIO: INTERNET oF THINGS FOR HEALTHCARE - gesture recognition
 \IEEEPARstart{I}{nternet}-of-Things is pervasively entering every field of
 society, including healthcare. Indeed, in the so-called
 \gls{ioth}~\cite{Islam2015}, several heterogeneous sensors, portable devices
 and wearables could be employed to continuously acquire large amounts of
 physiological data (e.g., heart rate, blood pressure and brain activity) as
 well as living context and environmental variables (e.g., artificial light
 exposure, air quality and (acoustic) noise level).
 Among other healthcare applications, gesture recognition can be evaluated by
 simultaneously recording brain and muscular activities via \gls{eeg} and
 \gls{emg}, respectively, and analyzing them to identify the specific movement
 that is being accomplished and the quality of its performance. It could be
 useful for assessing the advancements of training in patients recovering from
 neuro-motor diseases (e.g., stroke) or in athletes, where technical and complex
 gestures have to be optimally performed for a competition or recovered after an
 injury~\cite{Jovic2019}.
 However, wearing tens of sensors on the top of the head (to acquire brain
 activity) and along the limbs (to acquire muscular activity) can not be
 comfortable nor useful: unobtrusiveness and minimal setups are highly
 recommended in order not to interfere with the natural gestures and to ensure
 subject compliance. Moreover, an increasing number of IoT applications is being
 developed with the result that a growing amount of data is circulating the
 network, possibly leading to saturate the channel capacity and dramatically
 decrease the \gls{qos} of network users. Again, minimizing the acquisition
 setup and thus, the amount of data to acquire and process could represent a key
 solution to enabling a sustainable development of an IoT for healthcare
 (besides increasing the communication networks performance using new
 generations of communications technologies).
%
 % FEATURES SELECTION
 Therefore, in this paper we propose a robust feature selection algorithm,
 namely \gls{fesc} for \gls{eeg}/\gls{emg}-based gesture recognition (i) to
 robustly extract the most relevant features to classifying different grasping
 tasks, minimizing the acquisition setup to those \gls{eeg}/\gls{emg} sensors
 that only convey significant information and (ii) to retain the natural meaning
 of the selected features.

 The paper is organized as follows. In Sec.~\ref{sec:related-works} we present
 the related works; in Sec.~\ref{sec:preproc}, we describe how individual
 features are computed from EEG and EMG signals. Sec.~\ref{sec:pipeline}
 explains the details of the proposed feature selection algorithm.
 Sec.~\ref{sec:results} reports the results obtained during preliminary
 investigations on a publicly available dataset. The novelty of \gls{fesc} and
 its most significant differences and advantages compared to the state-of-art
 are discussed in Sec.~\ref{sec:discussion}. Finally, Sec.~\ref{sec:conclusions}
 concludes the paper, also mentioning possible extensions to this work.

 % RELATED WORKS
 \section{Related works}
 \label{sec:related-works}

 % Intro
 Feature selection is becoming critical in several applications of IoTH, e.g.,
 gesture recognition, emotion recognition and sleep/wake detection among
 others~\cite{csen2014comparative,Nakisa2018,Cisotto2018healthcom}.
 Particularly, finding a small number of highly discriminative features to
 compress the available big data could bring significant improvements in the
 classification performance~\cite{Khan2020, Cisotto2018globecom} and the
 reduction of the required communication resources.
%
 % [general feature selection studies]
 Feature selection has been explored in many
 studies~\cite{ienco2008exploration,liu2011feature,cai2018feature,galdi2018consensus},
 where several implementations of sequential forward/backward
 selection/elimination algorithms have been presented.
 % wahlberg2000methods, orhan2011eeg
%
 % hierarchical clustering
 Among other methods, hierarchical clustering was found to be a common choice to
 progressively reduce the number of features. In~\cite{ienco2008exploration},
 authors study the relevance of different feature combinations, thus providing
 an improvement of the accuracy performance using several
 classifiers. % in sec 3.5 they say it is a better approach than ranking the features
 \cite{liu2011feature} uses mutual information and the coefficient of relevancy
 to measure the distance between and within clusters,
 respectively. % increase of accuracy, as well?
 In~\cite{cai2018feature}, authors survey different methods for feature
 selection and describe several evaluation measures that can be used to compare
 their performance. Interestingly, they point out that clustering-based feature
 selection methods could retain irrelevant features: indeed, the latter are
 typically clustered together, forming large clusters which might be then
 represented by a representative feature(s) in the final set of selected
 features.
%
 % consensus clustering as solution
 As a solution,~\cite{galdi2018consensus} applies consensus clustering algorithm
 where multiple runs with different initializations of the clustering algorithm
 are combined together to return a smaller, but more robust, set of features.
 In~\cite{galdi2018consensus}, consensus clustering has been applied to
 neuroimaging signals, with the aim to reduce the spatial complexity of that
 dataset, while retaining the anatomical meaning of the
 signals. % They analyze different clustering and classification algorithms and see that random forest overcomes svm, and k-means is better than dbscan.
%
 % Relationship with current literature and novelty (1)
 In our work we apply a similar approach but, as input to the consensus
 algorithm, we take the outputs of two different clustering algorithms (i.e.,
 hierarchical and spectral clustering) and, as main novelty, we introduce
 consensus clustering to cluster \gls{eeg}/\gls{emg} features (themselves)
 together.
 % In this paper we follow a similar approach but, instead of applying the
 % consensus to multiple instances of the same clustering algorithm with
 % different initializations, we apply it to find the consensus among two
 % different clustering algorithms, i.e., hierarchical and spectral.
%
%
 % Preliminary "compression"/"data mining" based on physiology and novelty (2)
 Furthermore, we take advantage of the well-established physiological value of
 the \gls{msc}, a feature extracted from the joint analysis of \gls{eeg} and
 \gls{emg} signals.
 While the most literature on feature selection from heterogeneous signals
 (e.g., \gls{eeg}, \gls{emg}, \gls{eog}, \gls{ecg} and others) deals with
 unimodal features, i.e., extracted from one single kind of signal at a time, or
 artificial features, i.e., extracted using deep learning
 architectures~\cite{Said2017}, we consider a well-known multimodal feature that
 allows us to preliminary, and significantly, reduce the overall number of
 available features, while retaining, at the same time, their physiological
 relevance~\cite{Cisotto2018globecom}.
 Moreover, other previous works often limit the number of features by manually
 selecting, \textit{a-priori}, \gls{roi} (i.e., sensors) where features are
 extracted from. Here, instead, we use an extensive approach where all
 \gls{eeg} and \gls{emg} sensors are taken into account and selected manually.

 %% MSC COMPUTATION
 \section{Data pre-processing}\label{sec:preproc}
 In this section we describe the feature extraction and the dataset augmentation
 pre-processing steps.
 Given a pair of single-channel EEG and EMG signals, i.e., acquired from the
 \textit{h}-th (out of \textit{H}) EEG sensor and from the
 \textit{j}-th (out of \textit{J}) EMG sensor, respectively, let
 $s_{x}^{i}(t)$ and $s_{y}^{i}(t)$, with $i=1,2,....N$, represent all EEG and
 EMG segments, respectively, corresponding to $N$ repetitions (\textit{trials})
 of the gesture to investigate.
 Let $S_{x}^{i}(f)$ and $S_{y}^{i}(f)$, with $i=1,2,....N$, represent the
 autospectra of the \textit{i}-th segment of EEG and EMG, respectively, at
 repetition \textit{i}. Also, $S_{xy}^{i}(f)$ is the EEG-EMG (cross) power
 spectrum of the \textit{i}-th segment. The \gls{fft} algorithm is used to
 compute the spectra of segments.
 Then, $S_{x}(f)$, $S_{y}(f)$, and $S_{xy}(f)$ are the averages along the $N$
 repetitions, obtained from $S_{x}^{i}(f)$, $S_{y}^{i}(f)$, and $S_{xy}^{i}(f)$,
 respectively, with $i=1,2,...,N$.
 Finally, \gls{msc}$_{hj}$ \cite{MSC-NI} is the normalized $h$-th EEG-$j$-th EMG (cross) power
 spectrum, after averaging along repetitions, and it is obtained as follows:
 \begin{equation}
   MSC_{hj}(f)=\dfrac{|S_{xy}(f)|^2}{S_{x}(f)S_{y}(f)}
 \end{equation}
 Given the Cauchy-Schwarz inequality, it holds
 \begin{equation}
   0 \leq |S_{xy}(f)|^2 \leq S_{x}(f)S_{y}(f),
 \end{equation}
 and, as a consequence, $MSC_{hj}(f)$ assumes values between $0$ (uncorrelated
 signals) and $1$ (perfect linear relationship).
 From the \gls{msc} spectrum of each pair of EEG-EMG signals, we then extract the average \gls{msc} value
 of $K=11$ well-known different frequency bands: $\delta$ (1.5-4 Hz), $\theta$
 (4-8 Hz), $\alpha$ (8-13 Hz), $\beta_1$ (13-20 Hz), $\beta_2$ (20-30 Hz),
 $\beta$ (13-30 Hz), $\gamma_1$ (30-45 Hz), $\gamma_2$ (45-60 Hz), $\gamma_3$
 (60-80 Hz), $\gamma$ (30-80 Hz), and the full band (1.5-80 Hz). It is worth
 noting that some of them are overlapping with each other, e.g., $\beta_1$ and
 $\beta$, while others cover multiple narrower sub-bands, e.g., $\beta$ includes
 $\beta_1$ and $\beta_2$.
 All possible EEG-EMG-frequency band triplets, i.e., $\mathcal{M}=\{EEG_h,
 EMG_j, band_k\}$, with $h=1,2,....H$, $j=1,2,....J$, and $k=1,2,....K$ have
 been included into the feature matrix.
 The same feature extraction procedure is repeated for all repetitions of all types (i.e., classes) of motor task.
 Finally, to obtain balanced classes for a fair classification between different classes,
 we apply SMOTE to the smallest class \cite{smote2002}.
 Provided that a high-dimensional feature matrix is typically extracted at this
 step, an automatic procedure is needed to select the most relevant features for
 obtaining a reduced dataset (data mining) and to improving the classification
 performance.

 % %% PIPELINE
 \section{Proposed feature selection algorithm}\label{sec:pipeline}

 \begin{figure*}[t]
   \centering \includegraphics[width=\textwidth]{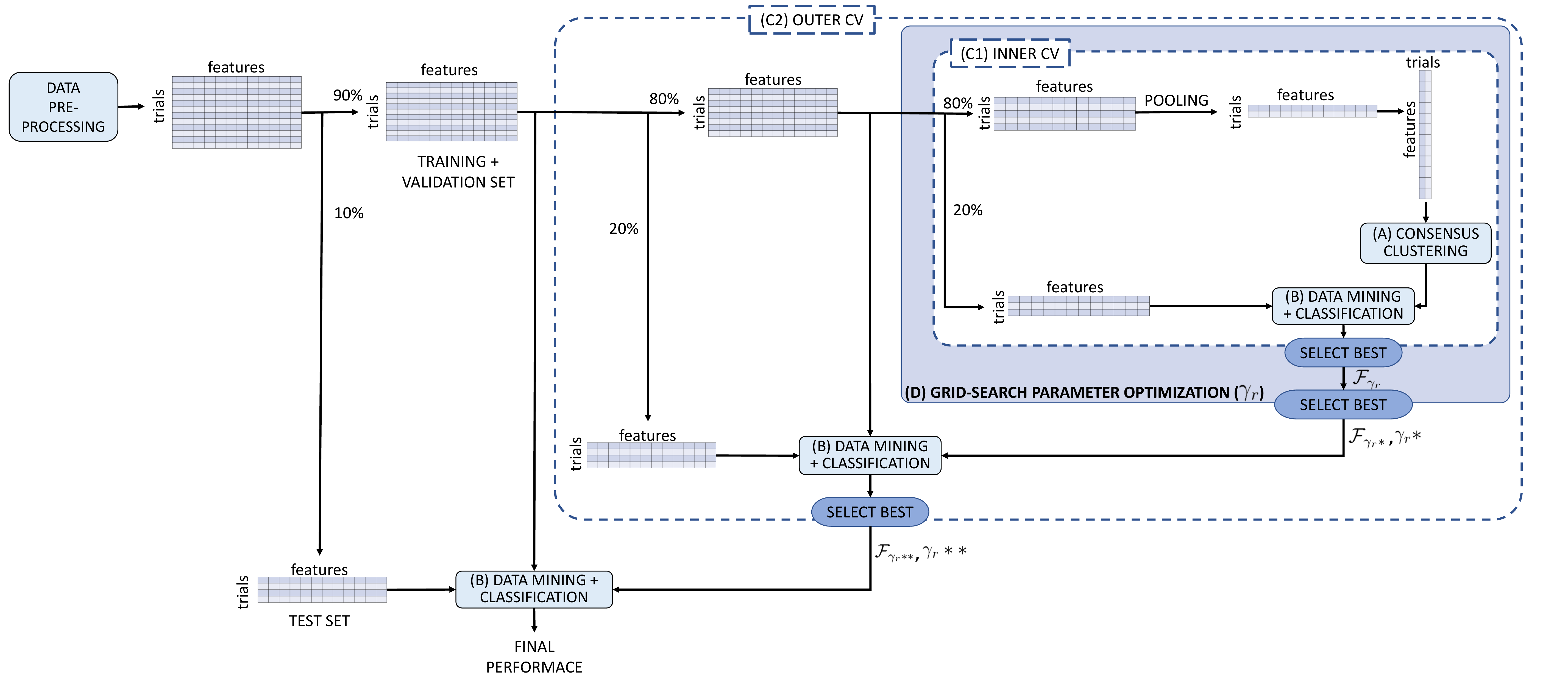}
   \caption{FeSC: the proposed feature selection algorithm. }
   \label{fig:pipeline}
 \end{figure*}

 % Target
 The aim of \gls{fesc} is two-fold: first, to
 reduce the number of features, keeping only the most informative ones using
 consensus clustering and, second, to select those which give the best
 performance in the classification of different grasping tasks.
 Nested \gls{cv} is implemented to realize a robust validation strategy~\cite{NestedCV2020}.
 To ensure a proper performance evaluation, $10\%$ of the dataset is kept for the final independent test,
 while the remaining $90\%$ is used for the nested procedure.
 Every partition used in \gls{fesc} includes the same proportion of class instances as the overall dataset.
 \gls{fesc} pipeline is represented in Fig.~\ref{fig:pipeline}: block $A$ represents the consensus clustering,
 block $B$ is for data mining and classification, blocks $C1$ and $C2$ include the \textit{inner} and
 \textit{outer} \gls{cv} loops, respectively, and block $D$ performs grid-search parameter optimization.
 In the following, we illustrate how the pipeline works, referring to Figs.~\ref{fig:pipeline},~\ref{fig:consensus}.

 %% FEATURE SELECTION
 \subsection{Consensus clustering}
 \label{sec:consensus}

 \begin{figure*}[t]
   \centering \includegraphics[width=1\linewidth]{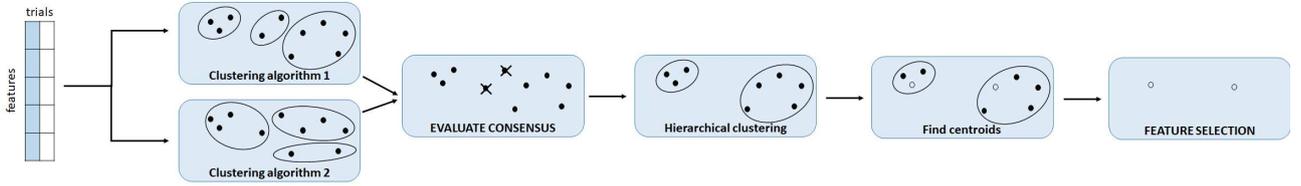}
   \caption{Consensus clustering (corresponding to block $A$).}
   \label{fig:consensus}
 \end{figure*}
 The algorithm for consensus clustering is represented by block $A$ in
 Fig.~\ref{fig:pipeline} and is further expanded in Fig.~\ref{fig:consensus},
 which can be used as a reference for the following
 description. % and it is executed for each of the folds during the \gls{cv}.
 We employ a consensus-based feature selection approach similar to what
 presented in~\cite{galdi2018consensus}. The input is defined as a vector
 $\gamma_r=[M_r, \sigma_r, \nu_r]$, with $r = 1, 2, ...R$ formed by the
 following clustering parameters: $M_r$, the maximum number of clusters on which
 the consensus algorithm works, i.e., that also corresponds to the number of
 features we decide to extract, and the thresholds $\sigma_r$ and $\nu_r$. The
 main steps for the consensus clustering are the following:
 \begin{enumerate}
 \item \label{mult_clust} Application of two clustering algorithms
   (agglomerative hierarchical and spectral clustering), separately, to the
   initial dataset;
 \item Evaluation of the consensus between them. This step is further divided
   into three substeps:
   \begin{enumerate}[i.]
   \item Computation of a pair-wise \textit{similarity matrix} for each
     clustering algorithm, indicating if two features are assigned to the same
     cluster.
   \item Computation of a \textit{consensus matrix}, which takes the average of
     the similarity matrices. In this matrix, the higher the value associated to
     a pair of features, the more often these features are clustered together by
     the different clustering algorithms.
   \item \textit{Consensus-based filtering}: application of thresholds
     ($\sigma_r$ and $\nu_r$) to the consensus matrix in order to select a
     subset of features (i.e., candidate to be finally selected). Given that
     $\sigma_r$ is related to the agreement between the different clustering
     algorithms, and $\nu_r$ is related to the average size of clusters, the
     features which have similarity greater than $\sigma_r$ with more than
     $\nu_r$ other features are only considered.
   \end{enumerate}
 \item \label{final_hier} Hierarchical clustering of features identified at the
   previous step, setting the number of clusters to $M_r$.
 \item Computation of the centroids of clusters found at the previous step:
   these are taken as the selected features.
 \end{enumerate}

 As mentioned before, we consider two algorithms for the consensus
 (step~\ref{mult_clust}): agglomerative hierarchical clustering and spectral
 clustering.

 \subsection*{Agglomerative hierachical clustering}
 Hierarchical clustering aims at grouping data over a variety of scales by
 creating a cluster tree called dendrogram~\cite{hastie2009unsupervised}. The
 dendrogram is a multilevel hierarchy in which clusters at one level are joined
 as clusters of the next level. In this way, it is possible to decide the most
 appropriate level of clustering.
 The hierarchical clustering algorithm works in the following way. First, it
 computes the distance between objects in the dataset based on a certain
 metric. %To compute the distance, we consider the
 % Euclidean distance between objects.
 % Then, objects are grouped into a binary hierarchical cluster tree by linking
 % pairs of objects that are in close proximity, determined through the distance
 % information previously generated.
 These distances are sorted in ascending order and objects with minimum distance
 are clustered together. Once objects are paired in binary clusters, these are
 paired in larger clusters. This procedure is repeated until all objects are
 linked together in a hierarchical tree. Finally, the hierarchical tree is cut
 in order to define the output clusters: to create a partition of the initial
 dataset, the branches at the bottom of the hierarchical tree are pruned off and
 all objects below each cut are assigned to a single cluster. In our
 application, we use Euclidean distance and Ward's method to group clusters; as
 cutting rule, we set the number of clusters to be equal to $M_r$.

 %% SPECTRAL CLUSTERING
 \subsection*{Spectral clustering}
 Spectral clustering was first introduced in~\cite{donath1973lower}, where
 authors proposed to make a partition in a graph based on eigenvectors (i.e., the
 ``spectrum'') of the similarity matrix of the data. In general, spectral
 clustering refers to a class of techniques relying on the eigenstructure of a
 similarity matrix to partition points into disjoint clusters: points having
 high similarity are assigned to the same cluster.
 %% DESCRIPTION OF SHI ALGO
 For our scope, we consider the spectral clustering algorithm presented
 in~\cite{shi2000normalized}, which employs the adjacency matrix, i.e., a matrix
 representing a graph, where elements indicate whether pairs of vertices are
 adjacent in the graph or not. The algorithm consists of two main steps:
 \begin{enumerate}
 \item Transformation of the adjacency matrix of the input data into a matrix
   with as many rows as the input elements and $M_r$ columns corresponding to
   the first $M_r$ eigenvectors.
 \item Application of the $k$-means algorithm~\cite{Bishop2006pattern} to
   clusterize the rows of the matrix above defined. Then, the output of the
   algorithm represents the $M_r$ clusters of the input elements.
 \end{enumerate}
 % It takes as input a similarity matrix $\mathbf{A}$ and the parameter
 % \textit{maxclust}, that in our case corresponds to the adjacency matrix and
 % to the desired number of clusters, respectively. Then, the unnormalized
 % Laplacian $\mathbf{L}$ is computed. At this point, the first
 % $k=$\textit{maxclust}
 % eigenvectors %$\mathbf{v}_1, \mathbf{v}_2,..., \mathbf{v}_k$
 % of the generalized eigenproblem $\mathbf{Lv}=\mathbf{\lambda D v}$ are
 % evaluated. The matrix $\mathbf{D}$ is obtained from $\mathbf{A}$ after adding
 % to it a few random edges to improve performance and multiply it by matrix
 % $\textbf{B}^{-1}$, where $\mathbf{B}$ is a diagonal matrix for which each
 % diagonal element $b_i$ is the sum of the $i$-th row of $\mathbf{A}$. Let
 % $\mathbf{V} \in \mathbb{R}^{n x k}$ be the matrix containing the eigenvectors
 % $\mathbf{v}_1,..., \mathbf{v}_{k}$ as columns, and let $y_i \in \mathbb{R}^k$
 % with $i=1,...,n$ be the vector corresponding to the $i$-th row of
 % $\mathbf{V}$. The core of the spectral clustering algorithm is to cluster the
 % points $(y_i)_{i=1,...,n}$ in $\mathbb{R}^{k}$ with the $k$-means algorithm
 % into clusters $\mathbf{C}_1,...,\mathbf{C}_k$. The output of the algorithm
 % are the clusters $\mathbf{E}_1,...,\mathbf{E}_k$ with $\mathbf{E}_i=\{ j |
 % y_j \in \mathbf{C}_i \}$
%
 %% GENERAL CONSIDERATIONS
 The success of spectral clustering is mainly based on the fact that no
 assumptions on the shape of clusters are needed. Moreover, it can be
 implemented efficiently even for large datasets, as long as the similarity
 graph is sparse. Indeed, once the similarity graph is chosen, the main step in
 the matrix transformation is the computation of eigenvectors which, being a
 linear problem, is amenable to simple and deterministic
 solution. % has no issues of getting
 % stuck in local minima, and there is no need of restarting the algorithm
 % several times with different initializations to find convergence.
 However, the choice of a good similarity graph is, in general, not
 trivial.% and spectral clustering could be quite unstable under different
 % choices of the parameters for the neighborhood graphs. As a
 % consequence, spectral clustering cannot serve as a black box
 % algorithm that automatically detects the correct clusters in any
 % given dataset. However, it can be considered as a powerful tool
 % producing extremely good results if applied with care.

 %% PERFORMANCE EVALUATION
 \subsection{Data mining and classification}
 \label{sec:validation}
 In these blocks we assess the classification performance over the current training set
 using the current selection of features.
 Therefore, we consider the validation set and apply data mining to retain only the
 features that are selected by the \textit{consensus clustering} procedure.
 Then, we classify heavy/light trials by applying \gls{svm}.
 \gls{svm} are models for supervised learning in (binary) classification and
 regression
 problems~\cite{Bishop2006pattern}. % In particular, during the first phase
 % (training), they take input elements labeled with two categories, and build a
 % non-probabilistic linear classifier to distinguish the two classes.
 The input elements can be represented as points in the feature space and the
 classifier as a hyperplane that is expected to optimally separate
 them. During the training phase, the \gls{svm} builds a non-probabilistic
 classifier to distinguish the two classes. In the second phase (test phase),
 new elements (belonging to the current validation set) are provided to
 the model and mapped in the same feature space:
 according to which side of the hyperplane they are located, they are classified
 into a class or the other. In our application, we consider an $M_r$-dimensional
 space and accuracy as main performance metric. Furthermore, with the
 \emph{kernel trick} it is possible to obtain nonlinear \gls{svm}; here, we
 compare the performance obtained using either a linear or a~\gls{rbf} kernel.

 %% CROSS_VALIDATION
 \subsection*{C1. Inner cross-validation}
 \label{sec:cvin}
 For the inner \gls{cv}, we use a 5-fold \gls{cv} (see Fig.~\ref{fig:pipeline}, block $C1$).
 % \textcolor{blue}{we use cv on the training+val set to select the best Mr,
 % that is the one that provides a higher accuracy/smaller MCE.}
 It consists in partitioning the dataset in 5 groups (folds), training the model
 on 4 folds, and finally testing it in the remaining fold. The training-test
 procedure is run 5 times, changing the fold that is used for validation at
 every iteration and keeping track of the evaluation performance.
 Before applying the consensus clustering procedure, we perform a pooling
 operation to reduce in a two-dimensional space the elements that characterize
 each feature: specifically, we take the mean of \gls{msc} across all trials of the
 same class ($\overline{MSC}_{heavy}$, $\overline{MSC}_{light}$).
 Therefore, each feature is now described by a vector $[\overline{MSC}_{heavy},
 \overline{MSC}_{light}]$. We apply the \textit{consensus clustering} as
 described in Sec.~\ref{sec:consensus}, and obtain one specific selection of features
 for each iteration of the inner \gls{cv}.
 Then, we evaluate the classification performance using a kernel-\gls{svm}
 classifier, as explained in Sec.~\ref{sec:validation}.
 Finally, we select the $\mathcal{F}_{\gamma_{r}}$ set of features which provides
 the best classification accuracy for the current combination of parameters $\gamma_{r}$.
 The procedure of this block is repeated for every combination of parameters.

 \subsection*{D. Grid-search parameter optimization}
 \label{sec:gridsearch}
 As sketched in Fig.~\ref{fig:pipeline}, grid-search parameter optimization is
 implemented to find the optimal combination of consensus clustering parameters
 that leads to the best classification performance with the minimum number of features.
 Therefore, the \textit{inner \gls{cv}} is repeated for every $\gamma_r$. We select the $\gamma_r*$ that achieves the best
 classification performance. If the same accuracy is obtained by two different inner folds
 (i.e., possibly two different selection of $M_r$ features), we retain that with
 the minimum number of selected features.
 The outputs of this block are $\gamma_r*$ and the set of best features $\mathcal{F}_{\gamma_{r}}^*$.

 \subsection*{C2. Outer CV}
 \label{sec:cvout}
  For the outer \gls{cv}, we use a 5-fold \gls{cv} (see Fig.~\ref{fig:pipeline}, block $C2$).
 % We observe that, due to the limited size of the dataset, the results obtained were dependent on the initial partition (i.e., which data were assigned to the training/validation datset).
 % Therefore, we introduced a \gls{cv} step which allowed us to select the best model (parameter combination and selection of features) across the different partitions.
% We used a 5-fold \gls{cv}.
 At each iteration, four folds are input to the grid-search parameter optimization (block $D$) and the inner
 \gls{cv} (block $C1$). The fifth fold is used as validation set to evaluate the model selected in block $D$.
 As previously (see Section~\ref{sec:cvin}), we select the fold associated with the highest accuracy.
 Therefore, the outputs of this block are $\gamma_r^{**}$ and the set of best features $\mathcal{F}_{\gamma_{r}}^{**}$.

 %% FINAL PERFORMANCE
 \subsection*{Final performance evaluation}
 \label{sec:final_perf}
 Then, we use the selection of features $\mathcal{F}_{\gamma_{r}^{**}}$ for the
 data mining step and the \gls{svm} training (block $B$), using the current training set
 (i.e., 90\% of the original dataset). Finally, we test this model to classify heavy/light
 trials in the independent test set (10\% of the original dataset).

 %% RESULTS
 \section{Results}\label{sec:results}

 % FIGURE 3 (training with different kernels)
 \begin{figure*}[t]
   \centering
   \begin{subfigure}[t]{\columnwidth}
     \includegraphics[width=0.95\columnwidth]{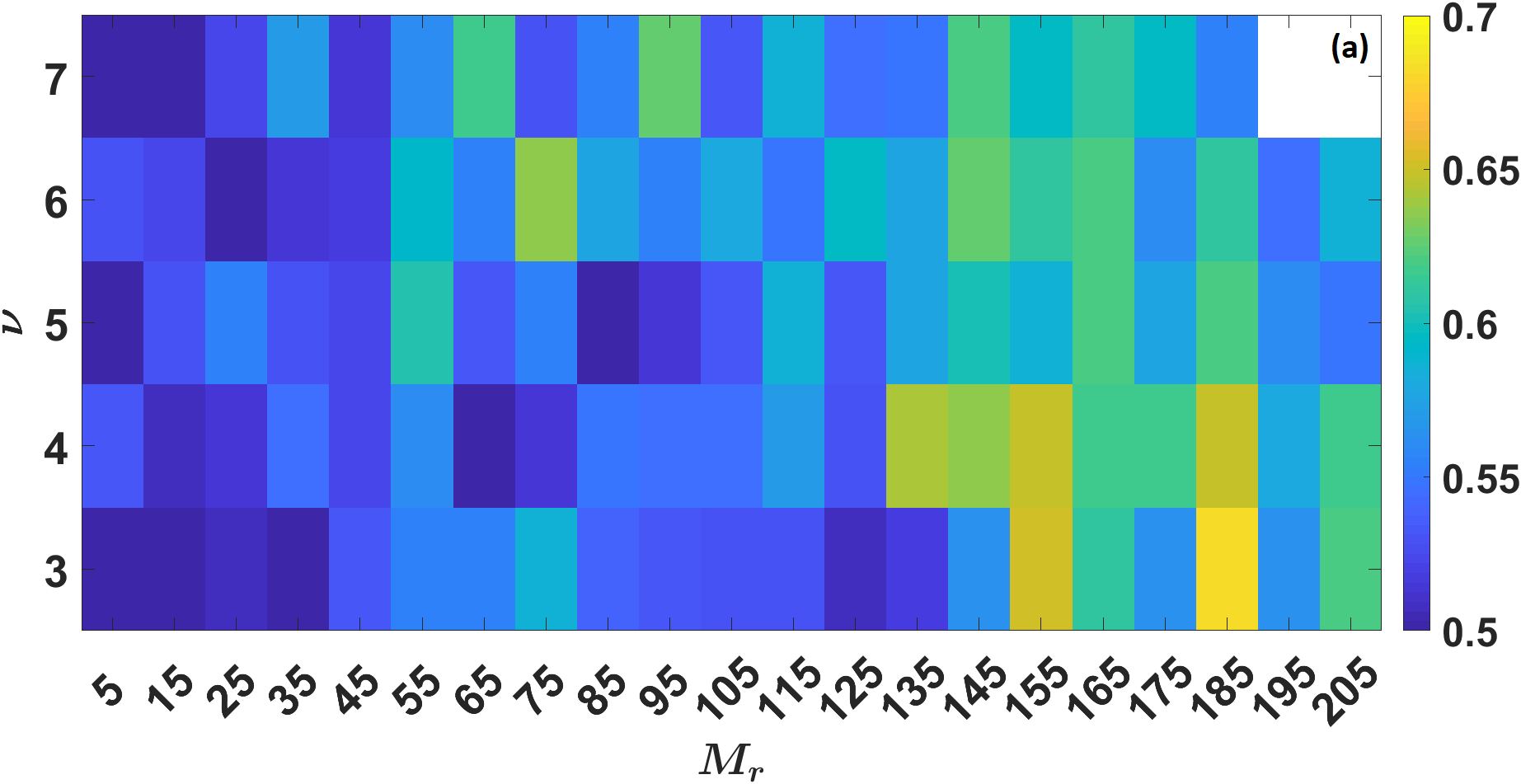}
     %\caption{Linear kernel.}
   \end{subfigure}%
   \hspace{2mm}
   \begin{subfigure}[t]{\columnwidth}
     \includegraphics[width=0.95\columnwidth]{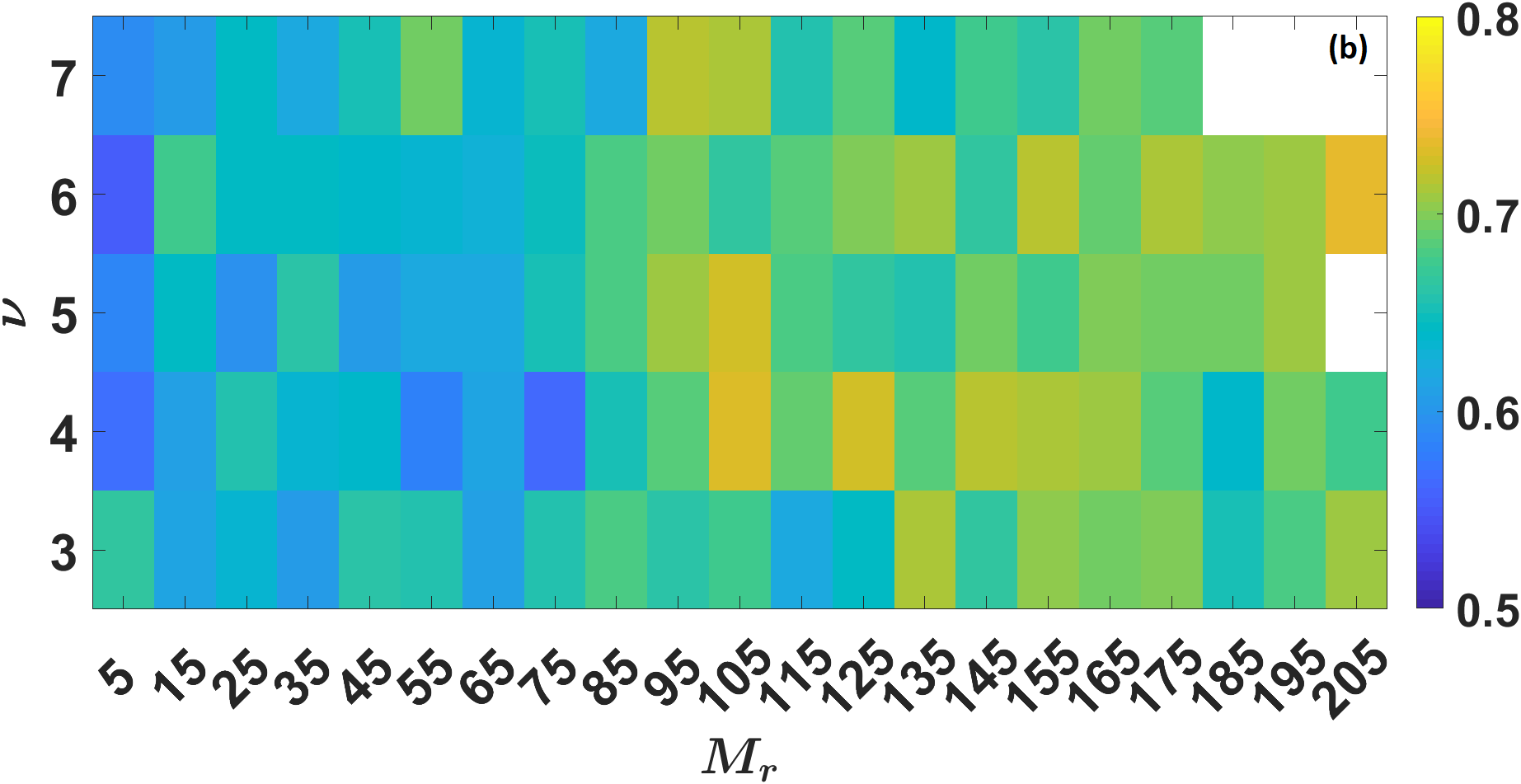}
     %\caption{Rbf kernel.}
   \end{subfigure}%

   \caption{Training performance in the inner CV fold giving $\gamma_{r}^{**}$
     and $\mathcal{F}_{\gamma_{r}^{**}}$ with variable number of features
     ($M_r$) and $\nu$ values. (a) Linear kernel. (b) Rbf kernel.}
   \label{fig:trainPerf}
 \end{figure*}

 \subsection{The dataset}
 The dataset is extracted from the large, publicly-available, WAY-EEG-GAL
 dataset~\cite{luciw2014multi}. EEG and EMG data (simultaneously recorded) were
 obtained while partecipants were performing repetitions (i.e., trials) of a
 grasp-and-lift task. At each repetition, they had to grasp an object with their
 thumb and index fingers, lift it up to an \textit{a-priori} selected position,
 hold it for a few seconds, and then return to the initial position, releasing
 the object.
 Different grasp-and-lift tasks were randomly proposed to the participants:
 the object could be changed in weight (i.e., light = 165 g, medium =
 330 g, and heavy = 660 g), surface friction (sandpaper, suede, and silk) or
 both. Here, we consider all repetitions where the object was light, as class
 $1$, and all repetitions where the object was heavy, as class $2$, regardless
 to their surface friction.
 Thirty-two EEG signals were acquired from thirty-two sensors, placed at
 standard locations (according to the International 10-20 EEG
 System). %\hl{ref in healthcom}).
 Moreover, ten EMG sensors were placed over five different muscles of the right
 upper limb, thus acquiring five bipolar EMG signals.
 Data from $8$ participants were included in this work (those without recodings issues).
 Then, we extracted 672 trials with light objects, 456 trials with heavy objects, 393 trials with sandpaper surface
 objects, and 1754 trials with silk surface objects. In this work, we classified light vs heavy trials.

 \subsection{Pre-processing}
 According to the experimental records available online, all data were
 artefacts-free. Therefore, we performed a lightweight pre-processing.
 Particularly, the EMG signals were downsampled to $500$~Hz (same sampling
 frequency as the EEG). Then, we used bandpass
 filtering (Chebyshev type I, order $86$) for EEG and EMG to limit the signals
 in the frequency band $1.5$-$80$~Hz. Moreover, we used notch filter (IIR
 filter, direct-form II, order $2$) to remove the power supply component in both
 signals. The filtered signals were segmented into $4$~s long segments, each
 one corresponding to a single repetition of the grasp-and-lift task. EMG segments
 were full-wave rectified. Then, all segments were normalized over their own
 area-under-curve (AUC) value. Finally, \gls{msc} was computed, as explained in
 Sec.~\ref{sec:preproc}, for each segment, resulting in a dataset of
 $5$x$32$x$11$ ($1760$) features with $456$ trials for heavy objects and $672$
 trials for light objects.
 As suggested by~\cite{smote2002}, we scale the input feature matrix and we applied
 SMOTE (see Sec.~\ref{sec:preproc}) to obtain balanced classes. Then, we obtain
 a normalized input feature matrix with 672 trials for both classes.
 \gls{fesc} pipeline has been fully implemented in Matlab and made available on GitHub\footnote{\url{https://signetlabdei.github.io/FeSC/}}.
%

% - - - FIGURE 4
% \begin{figure}[tpb]
%   \begin{center}
%     \includegraphics[width=0.9\columnwidth]{img2/selected_feat_CORE_v2.eps}
%     \caption{Feature selections (feature index on the y-axis) in the 5 outer CV folds (color code). Rbf kernel was used, $\nu$ set to 5 and $M_r$ values are reported on the x-axis. Due to space constraints we limit the visualization to the first \hl{30} features.} \label{fig:featselection} %\label{fig:featballs}
%   \end{center}
% \end{figure}
% - - -

 %% TRAINING PERFORMANCE
 \subsection{Parameters optimization and performance evaluation}

 \begin{table}[t]
   \centering
   \begin{tabular}[t]{cccc}
     \toprule
     \textbf{Kernel}       & \textbf{Parameters tuning}    & \textbf{Training} & \textbf{Final} \\
     \midrule
     Radial Basis Function & $[135,0.6,3]$                 & 0.6570            & 0.6045          \\
     \midrule
     Linear                & $[195,0.6,3]$                 & 0.5909            & 0.4851          \\
     \bottomrule
   \end{tabular}
   \caption{Training and test FeSC classification performance.}
   \label{tab:perf}
 \end{table}
 Fig.~\ref{fig:trainPerf} shows the training performance obtained in the inner CV
 (block $C1$) for all combinations of clustering parameters $\gamma_r =
 [M_r, \sigma_r, \nu_r]$ with two different kernel types, i.e., linear and rbf.
 Note that, since only two clustering algorithms have been used in the consensus
 clustering step, $\sigma_r$ is kept fixed to $0.6$ (i.e., to realize an actual consensus more
 than half clustering algorithms should agree on clustering two features together).
 Blank regions correspond to parameter combinations where consensus could not be found,
 while colored pixels represent reliable parameters combinations.
 It is worth noting that Fig.~\ref{fig:trainPerf} only reports results from one single
 inner \gls{cv} iteration, i.e., the one providing $\gamma_r*$ (for both kernels).
 Then, we obtained one such figure for each of the $5$ outer \gls{cv} iterations.
 However, here we report the one providing the best final parameters tuning.

 %% ---
 % \hl{prendere fold 4 per linear e fold 2 per rbf}
 %% ---
 %
 %
 Note that, at each iteration of the outer \gls{cv} loop, we can possibly find a different $\gamma_r*$, i.e., $M_r*$, and
 $\mathcal{F}_r*$, i.e., specific combination of features (even in case of same $M_r*$ value).
%
 % We have one of these plot for each of the 5 folds of the outer CV, but here we report only that for the best fold selected in the outer CV (block $C2$).
 We can observe that, for a fixed $\nu_r$, the accuracy increases with the number $M_r$ of features,
 regardless to the kernel choice. However, a significant improvement is obtained using an rbf kernel.
 Therefore, in the following, we consider this kernel, only.
 From Fig.~\ref{fig:trainPerf}(b) we can see that the best parameters tuning in output from the
 inner \gls{cv} (block $C1$) corresponding to one out of $5$ outer \gls{cv} iterations
 was $\gamma^* = [205, 0.6, 6]$, which achieved an accuracy of $0.7358$.
 % fullCOSTtrain(5,21,4,4)
 Then, those 205 features were then used for data mining and classification during the corresponding outer \gls{cv} iteration.
 Finally, the combination of features $\mathcal{F}_r^{**}$ and parameters tuning $\gamma_r^{**}$ which provided the best
 accuracy across all outer \gls{cv} iterations were selected as the best model for our classification problem (see Section~\ref{sec:validation}).
 Therefore, the overall training set was used for training a kernel-\gls{svm} classifier with the $M_r^{**}$
 $\mathcal{F}_r^{**}$ features. Then, the independent (unseen) test set was used to evaluate \gls{fesc} performance.
 Tab.~\ref{tab:perf} summarizes these results.
 Incidentally, we notice that the best model ($\gamma_r^{**}$, $\mathcal{F}_r^{**}$) does not match the one reported in Fig.~\ref{fig:trainPerf}. In fact, the $\gamma_r$ and $M_r$ features selected at the output of one inner \gls{cv} inner loop might not represent the best model across all outer \gls{cv} iterations.

 % As an example, Fig.~\ref{fig:featselection} reports the different selections of features obtained in the 5 outer CV folds.

 % As it can be observed, they may differ consistently. In fact, we found that
 % no features are common to all folds, while pairs of folds may share few
 % features.

 %% DISCUSSION
 \section{Discussion}\label{sec:discussion}
 % Summary of our work
   In this work, we propose \gls{fesc}, a new feature selection algorithm for \gls{ioth}
   applications, e.g., gesture recognition based on EEG and EMG signals.
   FeSC is a two-step processing pipeline including: (1) a consensus clustering step to identify
   the most relevant features to classify two different gestures, e.g., grasping of heavy/light
   objects, and (2) a data mining step to reduce the dataset for improving classification.
   %
%  Recap of results and proof that we fullfil objectives of the work
   We show that our algorithm can classify heavy/light objects using a limited number of features (much smaller than those available in the high-dimensional input feature matrix).
   We also observe that a proper choice of the \gls{svm}'s kernel can further improve the
   classification performance (as reported in Tab.~\ref{tab:perf}).
   % Particularly, the rbf kernel brings to a significant reduction of the number of features needed to have accuracies above \hl{$0.9$}(i.e., $M_r = 195$ is needed for linear kernel, while $M_r = 135$ is enough in case of rbf kernel).
   %
% ROBUSTNESS
   Robustness of our results is ensured by the consensus clustering step which selects only those features
   relevant to all (i.e., two) clustering algorithms and by the nested cross-validation used for validation.
   At the same time, differently from other approaches (e.g., autoencoders~\cite{Said2017}), \gls{fesc} makes
   it possible to preserve the physiological meaning of features to allow interpretation by clinical experts.
   %
 % Novelty and advantages
   Taking into account only those \gls{eeg}/\gls{emg} sensors leading to relevant features, we could
   propose a minimal \gls{eeg}/\gls{emg} setup for gesture recognition. This represents a major advantage
   in this kind of IoTH applications, where individuals are typically more compliant to minimally intrusive
   sensing platforms.
   Moreover, limiting the number of sensors for accurate classification of different gestures allows to
   reduce the amount of data to be stored or transmitted over the \gls{ioth} network, with possible savings of, e.g.,
   wireless sensors' batteries, network channel bandwidth, \textit{Cloud} resources and computational power to
   instantaneously analyze data.
   %
 % Limitations and future works
   Despite of \gls{fesc}'s robustness, our work still suffer from some limitations: the most relevant one is
   the selection of the specific features to use in the classification. At present, we select the combination
   of $\{EEG, EMG, band\}$ features which shows the best classification accuracy across folds, both in the inner
   and in the outer \gls{cv} loops, with the same approach proposed by~\cite{NestedCV2020}.
   However, different strategies could be employed either by selecting a subset of common features across the
   folds of the inner \gls{cv} loop~\cite{NestedCV2020}, or by adding some pre-processing steps to aggregate
   data in the spatial, i.e., sensors, dimension~\cite{galdi2018consensus}.
   Moreover, a larger and more variable dataset could help to gain higher classification performance (preliminary observations over larger datasets, not reported here, shown promising outcomes that need to be confirmed with further investigations).
   Then, different input features (e.g., partial directed coherence) and additional clustering algorithms for the consensus could be tested to improve classification performance.
   Finally, the application of \gls{fesc} on different datasets, e.g., including different gestures~\cite{Cisotto2018healthcom}, could provide the generalization capabilities of \gls{fesc} and can help identifying a common minimal recording setup to use in a variety of out-of-the-lab \gls{ioth} applications sharing the same classification target.

 %% CONCLUSIONS
 \section{Conclusions}\label{sec:conclusions}
 This work presents \gls{fesc}, a new and robust feature selection algorithm to
 extract the most relevant $\{EEG, EMG, band\}$ features for classifying
 different gestures, i.e., grasping of different objects, while retaining their
 physiological meaning.
 \gls{fesc} implements a very robust architecture for feature selection and classification.
 However, our results show that a limited number of features struggles to reach good classification
 performance, probably due to the limited size of the dataset.
 In the future, further investigations could determine to what extent size could cause a drop in the performance
 of \gls{fesc}.
 Moreover, we could extend this work to other gestures and, possibly, biosignals in order to extend
 the applicability of our classification algorithm to new \gls{ioth} applications.

% Therefore, as next steps, we plan to overcome this problem by considering other approaches to better calibrate the feature selection step (e.g., adding spatial information or multiple clustering algorithms), and by employing it to larger and different dataset, to increase both its performance and extend the range of applicability range to other \gls{ioth} applications.

 % ACKNOWLEDGMENT
 \section*{Acknowledgments}
 \small{The authors would like to thank Prof. Giorgio Maria Di Nunzio, Dept.
   Information Engineering, University of Padova, for his support in the
   pipeline design.

   Part of this work was supported by MIUR (Italian Minister
   for Education) under the initiative "Departments of Excellence" (Law
   232/2016).}

 \bibliography{biblio} \bibliographystyle{IEEEtran}

\end{document}